\documentclass{article}

\usepackage{graphicx} 
\usepackage[utf8]{inputenc}
\usepackage{algorithm}
\usepackage{algpseudocode}
 \usepackage{float}
\usepackage{graphicx}
\usepackage{glossaries}
\usepackage{tikz}
\usepackage{placeins} 
\usepackage{titling}
\usetikzlibrary{shapes.geometric, arrows, positioning}

\usepackage{glossaries}  
\usepackage{hyperref}
\makeglossaries   

\newacronym{hvac}{HVAC}{Heating, Ventilation, and Air Conditioning}
\newacronym{rl}{RL}{Reinforcement Learning}
\newacronym{ml}{ML}{Machine Learning}
\newacronym{drl}{DRL}{Deep Reinforcement Learning}
\newacronym{mfrl}{MFRL}{Model Free Reinforcement Learning}
\newacronym{mbrl}{MBRL}{Model Based Reinforcement Learning}
\newacronym{sac}{SAC}{Soft Actor Critic}
\newacronym{tl}{TL}{Transfer Learning}
\newacronym{cl}{CL}{Continual Learning}
\newacronym{boptest}{BOPTEST}{Building Optimization Testing}  
\newacronym{bhhp}{BHHP}{Boptest Hydronic Heat Pump}  
\newacronym{mpc}{MPC}{Model Predictive Control}  
\newacronym{nlp}{NLP}{Natural Language Processing}  
\newacronym{cv}{CV}{Computer Vision}

\title{Continual Reinforcement Learning for HVAC Systems Control:\\ Integrating Hypernetworks and Transfer Learning}

\author{
  Gautham Udayakumar Bekal \\
  Mitchell, Enlyte \\
  \texttt{gauthambekal93@gmail.com}
  \and
  Ahmed Ghareeb \\
  Department of Mechanical Engineering \\
  College of Engineering, University of Kirkuk \\
  \texttt{aghareeb@uokirkuk.edu.iq}
  \and
  Ashish Pujari \\
  Department of Mechanical Engineering \\
  University of North Carolina at Charlotte \\
  \texttt{apujari1@charlotte.edu}
}

\date{March 2025}

\begin{document}
\maketitle

\section*{Abstract}
Buildings with \gls{hvac} systems play a crucial role in ensuring indoor comfort and efficiency. While traditionally governed by physics-based models, the emergence of big data has enabled data-driven methods like \gls{drl}. However, \gls{rl}-based techniques often suffer from sample inefficiency and limited generalization, especially across varying \gls{hvac} systems.

We introduce a model-based reinforcement learning framework that uses a Hypernetwork to continuously learn environment dynamics across tasks with different action spaces. This enables efficient synthetic rollout generation and improved sample usage.

Our approach demonstrates strong backward transfer in a continual learning setting: after training on a second task, minimal fine-tuning on the first task allows rapid convergence within just 5 episodes—outperforming \gls{mfrl} and effectively mitigating catastrophic forgetting.

These findings have significant implications for reducing energy consumption and operational costs in building management, thus supporting global sustainability goals. \\

Keywords: Deep Reinforcement Learning; HVAC Systems Control; Hypernetworks; Transfer and Continual Learning; Catastrophic Forgetting 

\clearpage

\subsection*{Nomenclature}
\begin{description}
  \item[HVAC] Heating, Ventilation, and Air Conditioning
  \item[DRL] Deep Reinforcement Learning
  \item[RL] Reinforcement Learning
  \item[TL] Transfer Learning
  \item[CL] Continual Learning
  \item[MFRL] Model-Free Reinforcement Learning
  \item[MBRL] Model-Based Reinforcement Learning
  \item[SAC] Soft Actor Critic
  \item[MPC] Model Predictive Control
  \item[BOPTEST] Building Optimization Testing Framework
  \item[BHHP] BOPTEST Hydronic Heat Pump
  \item[ML] Machine Learning
  \item[NN] Neural Network
  \item[RL Agent] Reinforcement Learning Agent (Actor-Critic or Q-learning based)
  \item[Hypernet] Hypernetwork (generates weights for a target network)
  \item[Catastrophic Forgetting] Rapid loss of previously learned knowledge when adapting to new tasks
  \item[Dyna] Integrated architecture combining learning, planning, and reacting
\end{description}

\subsection*{Symbols and Notation}
\begin{description}
  \item[\(\gamma\)] Discount factor for future rewards, \(0 < \gamma \le 1\)
  \item[\(\alpha\)] Learning rate or regularization coefficient (context-dependent)
  \item[\(\beta\)] Regularization term coefficient used in hypernetwork training
  \item[\(s\)] Environment state (for example, zone temperature or time)
  \item[\(a\)] Action selected by the RL agent (such as a discretized control input)
  \item[\(r\)] Reward returned by the environment for a given \(s\)-\(a\) pair
  \item[\(s'\)] Next state resulting from the transition \((s,a)\)
  \item[\(\pi_\theta\)] Policy network with trainable parameters \(\theta\)
  \item[\(Q_\phi\)] State-action value (critic) network with parameters \(\phi\)
  \item[\(H_\varphi\)] Hypernetwork with parameters \(\varphi\)
  \item[\(T_\delta\)] Target network (e.g., a dynamics or reward model) generated by \(H_\varphi\), with parameters \(\delta\)
  \item[\(\mathcal{M}_\alpha\)] Real data memory buffer
  \item[\(\mathcal{M}_\beta\)] Synthetic data buffer (model-generated samples)
  \item[\(\mathcal{M}_\gamma\)] Hypernetwork training data buffer
  \item[\(K_i\)] Number of training episodes or iterations for task \(i\)
  \item[\(\text{task-id}\)] One-hot identifier specifying the current task
  \item[\(\text{layer-id}\)] Layer index used by \(H_\varphi\) to generate target network parameters
  \item[\(P(s' \mid s,a)\)] Transition dynamics (approximated by the hypernetwork)
  \item[\(\eta\)] Learning rate for the SAC or hypernetwork optimizer
\end{description}

\section{Introduction}

Heating, Ventilation, and Air Conditioning (HVAC)  systems have a major impact on energy use and consumption and are essential to maintaining indoor comfort, air quality, and overall building function. With increased efficiency, they have the potential to save 20–40\% of the total energy consumed in commercial buildings and up to 48\% in residential buildings. In a particular environment, HVAC systems can account for 60–70\% of a building energy demand, resulting in increased operating costs and greenhouse gas emissions. In addition to lowering costs and having a positive environmental impact, increased HVAC efficiency also increases occupant productivity and comfort. Innovative control strategies, such as transfer learning (TL) and reinforcement learning (RL), present viable options to effectively and sustainably maximize HVAC performance and energy efficiency.

Since the advent of deep learning a variety of \gls{rl} based approaches have been developed to solve HVAC control-related issues. However, RL-based approaches primarily have two challenges, sample inefficiency and generalizability. 
One of the ways to handle sample inefficiency is using model-based reinforcement learning ~\cite{Janner} which involves explicitly creating an approximate model of the environment and letting the RL system learn from both the actual and the surrogate environment. The second problem of generalization is usually dealt with using the transfer learning approach, which has been widely used in natural language processing and computer vision ~\cite{Zhuang}.
In the case of \gls{rl} for HVAC systems, this involves training an agent in a specific HVAC simulator and then varying the building parameters during the fine-tuning phase.
Unfortunately, such approaches have a severe shortcoming in the \gls{ml}  and \gls{rl} domains, since the parameters of the neural network are optimized to minimize the loss on the latest task at hand, which means the model will perform extremely poorly in the older tasks.
This problem is well documented in machine learning literature and is called the Catastrophic Forgetting problem ~\cite{Luo}

In order to mitigate both of the above issues, we combine \gls{sac} based \gls{rl} algorithm with Hypernetworks. Crucially, the \gls{sac} utilizes vanilla neural nets, and the model of the environment is learnt using Hypernetworks.
We carry out continual transfer learning using various action spaces across the tasks and the Hypernets continually learn these different configurations of the environment.
The hypernet then augments the data set of real samples with synthetic data samples to speed up the training of \gls{rl} agent under Dyna-style ~\cite{Sutton} model-based RL
~\cite{Moerland}.

\section {Related Work}

\gls{drl} has been proven to be a powerful technique in traditional ML experiments such as games ~\cite{Mnih}. Its applications have now been impacted in other domains such as tools for optimizing control systems, particularly in HVAC applications ~\cite{Sayeda}. Traditional control methods, such as \gls{mpc} ~\cite{Taheri} and other classical approaches, often require extensive manual tuning and may not adapt easily to changing conditions or uncertainties. It usually struggles with efficiency and adaptability, prompting researchers to explore the potential of RL to enhance energy conservation and indoor comfort. In this context,  Gao and Wang ~\cite{Gao} investigate HVAC system RL techniques that are model-based and model-free. They compare the computing needs and efficiency of both methods using the \gls{boptest} framework. Their implementation combines model-based techniques and sophisticated \gls{rl} methods such as Dueling Deep Q-Networks and Soft Actor-Critic. Comparing RL controllers to conventional approaches, they find that all of the approaches improve interior temperature regulation and save operating costs. Remarkably, despite initial model flaws, model-based reinforcement learning outperforms model-free RL with shorter training times. 

More recently, \gls{tl} has been extensively studied across various domains but has yet to be thoroughly investigated in the HVAC systems sector.  For instance, a unique \gls{tl} framework is proposed by Coraci et. al ~\cite{Coraci} in order to enhance the scalability of \gls{drl} controllers in buildings that include integrated energy systems. The authors introduced a heterogeneous \gls{tl} technique to overcome the scalability and generalization limitations of \gls{drl}. This approach is tested in Energyplus simulations with a range of target buildings, concentrating on a \gls{drl} policy for controlling a chiller in conjunction with thermal energy storage. Model slicing, imitation learning, and fine-tuning are used in the \gls{tl} technique to address a variety of building situations and characteristics. The proposed method effectiveness and applicability in the test environments demonstrated that it lowers costs of electricity and temperature violation rates while simultaneously improving self-sufficiency and self-consumption ~\cite{Maier}.

In prior work, researchers explored techniques to accelerate DRL for energy management by leveraging learned hidden layers from a pretraining phase for fine-tuning. Since states and actions can vary across environments, some re-engineering is required.

While the presented studies show the effectiveness of \gls{tl} in enhancing HVAC control, they lack extensive investigation into \gls{cl} setup where the system needs to perform well on both forward and backward transfer. Additionally, the literature often presents transfer learning across changing state space, whereas here we present transfer learning across action space. We also observe that despite presenting valuable RL frameworks, crucial parts are sometimes missing, making it difficult to reproduce, validate, or build upon the work.

This study fills a significant gap by presenting a novel approach to continual learning \gls{rl} through the use of an environment model, which improves generalization.
It provides a comprehensive framework for transferring knowledge between environment configurations differing in action space and thus, addressing limitations in existing literature, and enhancing the effectiveness of the RL technique.

Kadamala et. al ~\cite{Kadamala} proposed a modular neural net architecture where the 
parameters for the hidden layers are transferred from the pre-training environment
to the fine-tuning environment. The assumption is that the RL agent trained on the first environment should have information useful for the second environment, which is stored in the parameters.
However, they were only able to get a marginal improvement of 1 to 4\%. This confirms our analysis that transfer learning of direct RL agents leads to minimal improvement.

Xu et. al ~\cite{Xu} used a different neural net architecture to carry out
transfer learning across similar environments. However, akin to the aforementioned study,
we see that transfer learning is at the RL agent level instead of the environment
level.

Gao and Wang ~\cite{Gao} analyzed model-based and model-free RL using
actor-critic and Q-learning approaches. Here, they show the sample efficiency
problem being mitigated compared to the vanilla version of the RL algorithm.
However, the authors carry out only a single experiment. We extend their
approach by carrying out detailed experiments for two environments and also
analyze the backward transfer performance by setting up continual learning
setup.

Our approach is quite generic and even though we have used \gls{sac} in our experiments it can be modified to handle any standard \gls{rl} algorithm.

Catastrophic forgetting has been extensively studied in \gls{ml} literature and we point the interested reader to go through them~\cite{Fernando}. 
Crucially, the problem of Catastrophic forgetting is a tradeoff between the performance of forward task and backward task in a continual learning setup.

The following points highlight our contribution to the literature:

\begin{itemize}

    \item By using Hypernets for learning the environment models, we create a world model, that acts as a substitute for the real environment and learns the dynamics of environment configurations differing in action spaces.

    \item We carry out \gls{cl} at the environment model level and thus utilize MBRL to improve sample efficiency across tasks.
        
    \item Using the regularization term in the training of Hyperent, we significantly reduce the problem of catastrophic forgetting.
\end{itemize}

\section{Preliminaries}

In this section, we review the key concepts that form the foundation of our work. Each concept is introduced in a way that motivates the need for the next, creating a logical progression toward the core ideas of our approach.

\subsection{\gls{boptest} Simulator}
\gls{boptest} \cite{Blum} is a standardized simulation framework for evaluating advanced control algorithms in building energy systems. It provides a set of virtual building models with realistic HVAC dynamics, weather data, internal gains, and occupant behavior.
\gls{boptest} has several use cases to choose from, and for this work, we choose \gls{bhhp} since this has been utilized in the existing literature and thus is easier to compare to.
For our work, we choose the time, current room temperature, and dry bulb temperature forecast, as input states to SAC.
We also add cooling and heating setpoints as inputs to the Hypernetwork and discard time to improve reward prediction. Adding setpoints to SAC had minimal impact hence discarded from SAC.
We change the action space as per the task at hand we are training the model on.
We define the collection of action 1 as the heat pump modulating signal, action 2 as the heat pump evaporator fan, and action 3 as the emission circuit pump as the action space.
Here, actions used to control the environment temperature are discrete, and hence actions generated by the \gls{rl} agent need to be discretized. 
The reward is generated by the BOPTEST environment and it involves reducing the thermal discomfort of the occupants in the room.
You can refer to the tables ~\ref{tab:hypernet-io}, ~\ref{tab:targetnet-io} and ~\ref{tab:rl-agent-io} for more detals.

\subsection{Soft Actor-Critic}
Soft actor-critic is a state-of-the-art \gls{rl} which directly optimizes the policy to maximize long-term discounted rewards ~\cite{Haarnoja}
The actor takes the state as input and generates action to maximize reward, and there is a trade-off between exploration and exploitation. The critic uses bootstrapping to estimate the state-action value Q(s,a) which the actor takes as input.
Concretely to train \gls{sac} we need tuples of (state, action, next-state, reward).

\subsection{Model Based Reinforcement Learning}
The SAC algorithm described above relies on sampling tuples of (s, a, s', r) for training. However, if we had a statistical model such as a neural net that has learnt to predict s' and r given s, a as inputs then the \gls{rl} algorithm can query the environment to obtain these generated tuples and train the model. However, the core issue is the predicted s' and r should be accurate and unbiased else it can destabilize the \gls{rl} algorithm.
Hence, in \gls{mbrl} we train both an RL agent and a model of the environment simultaneously which greatly improves sample efficiency.
In our use case we use Hypernets \cite{Chauhan} to learn the dynamics of the environment and predict the next state and reward. We follow the Dyna-style model-based RL framework here.

\subsection{ Transfer Learning}
\gls{tl} involves two stages of training, where in stage 1 we train on task 1 and in stage 2 we train on task 2, using the pre-trained weights after task 1 as the initial weights for the model. 
This helps in speeding up the training process for task 2. \gls{tl} has been utilized successfully in both \gls{nlp} and \gls{cv}.

\subsection{ Continual Learning}
Unlike in \gls{tl}, in \gls{cl} \cite{Wang} we train on multiple tasks in a sequential fashion. Crucially, the difference is in \gls{cl} we expect the learnt model to perform well on all the tasks and not just on the latest fine-tuned task as in \gls{tl}. This makes \gls{cl} much more difficult since both forward and backward transfer performance needs to be satisfactory.
Thus a major theme in \gls{cl} is to mitigate catastrophic forgetting in older tasks, and thus begs the need of sophisticated techniques such as ~\cite{Rolnick}.

\subsection{Hypernetworks}

A Hypernetwork \cite{Chauhan} is a neural network that generates the weights of another neural network. Formally, given a Hypernetwork \( H_\phi \) parameterized by \( \phi \), and a target network \( f_\theta \), the Hypernetwork produces \( \theta = H_\phi(z) \) where \( z \) is an input (e.g., a task embedding). This allows parameter generation conditioned on task-specific or context-specific information.
In our experiments, we generate task-specific embedding as one hot encoded value and add Gaussian noise to generate an ensemble of target weights. This acts similarly to Bayesian neural networks and reduces systemic bias which can severely impact the performance of model-based \gls{rl}.

\section{Methodology}
We train our model based \gls{rl} in Dyna-style fashion. Here, we simultaneously train both the SAC-based \gls{rl} agent along with Hypernets at each time step.
Here the SAC-based \gls{rl} agent learns to maximize the rewards, whereas the hypernet is learning the dynamics of the environment.
The \gls{rl} agent uses both real and synthetic data for training. Initially, the hypernet predictions are poor but after some time steps, the model starts generating high-quality predictions which the \gls{rl} agent consumes.

The following section provides a detailed explanation of the hypernetwork training procedure.

\subsection{Task Setup}
In our continual learning setup, we have 3 tasks.
Task 1 uses only a single action ( Heat pump modulating signal ) as the control input to the \gls{bhhp} environment. Here, the other two actions ( heat pump evaporator, emission circuit pump) to control the room temperature are in default setting generated by \gls{bhhp} itself.

In Task 2, the actor generates all 3 actions output to have a more fine-grained control of the room temperature.

In Task 3 the actor is in a similar setup as in Task 1, generating only a Heat pump modulating signal, for controlling the environment and the remaining two actions are the default values from 
\gls{bhhp}.

The Hypernetwork always receives all three actions along with the state as input and generates the next state and reward as output. The specific combination of actions—i.e., whether two actions are produced by the actor policy or are default values from the BOPTEST environment—varies depending on the task. This variation in action configuration across tasks establishes a transfer learning setup, where the Hypernetwork must generalize across different control schemes.

\subsection{Training Soft Actor Critic}
For all three stages, we train soft actor-critic from scratch. 
Readers can go through ~\cite{Haarnoja} for more details. The difference is, in \gls{mbrl} we train the SAC algorithm using both real and synthetic samples of (s, a, s', r).

\subsection{ Training Hypernetworks}

The Hypernet takes in task-id and layer-id as input and generates parameters for the target network. The forward pass of the target network is same as that of a regular neural net and takes in current state and action as input and generates next sate or reward as the output.
Specifically, in our experiments we have two pairs of Hypernet-target-network setups. The first one takes the state-action pair as input and generates the next state as the output. 
The other one takes in the same state action as the input and generates a reward as the output.
Gaussian noise is added to the task-id before feeding to Hypernet, so that Hypernet generates a distribution of target parameters instead of a single parameter, thus reducing bias which may cripple the \gls{rl} agent.
The Hypernet generates the parameters of the target network on a layer-by-layer basis which reduces the parameters needed to train the Hypernet.
It is important to remember that only the Hypernet is trained using backpropagation. The target network is only used to generate the predictions at the output.

\subsection{Mitigating Catastrophic Forgetting in Hypernets}

Among various methodologies to mitigate catastrophic forgetting in neural nets, Hypernets \cite{Oswald} offer superior solutions in terms of performance and memory consumption. 
In our scenario, Hypernets will dynamically generate parameters conditioned on task-id and layer-id.
This means that Hypernets are trained on task 2, it will rewrite the weights of task 1 due to catastrophic forgetting, which leads to performance degradation of task 1. 
To prevent this, Hypernet is trained with mean square error loss along with a regularization term. Concretely, the Hypernet generates the parameters of all the previous tasks  along with the parameters of the current task, which means that the model balances in improving the performance on the current task (forward transfer) as well as not degrading the performance of previous tasks (backward transfer)

\vspace{1em} 

\begin{figure}[H]
  \centering
  \includegraphics[width=\linewidth,keepaspectratio]{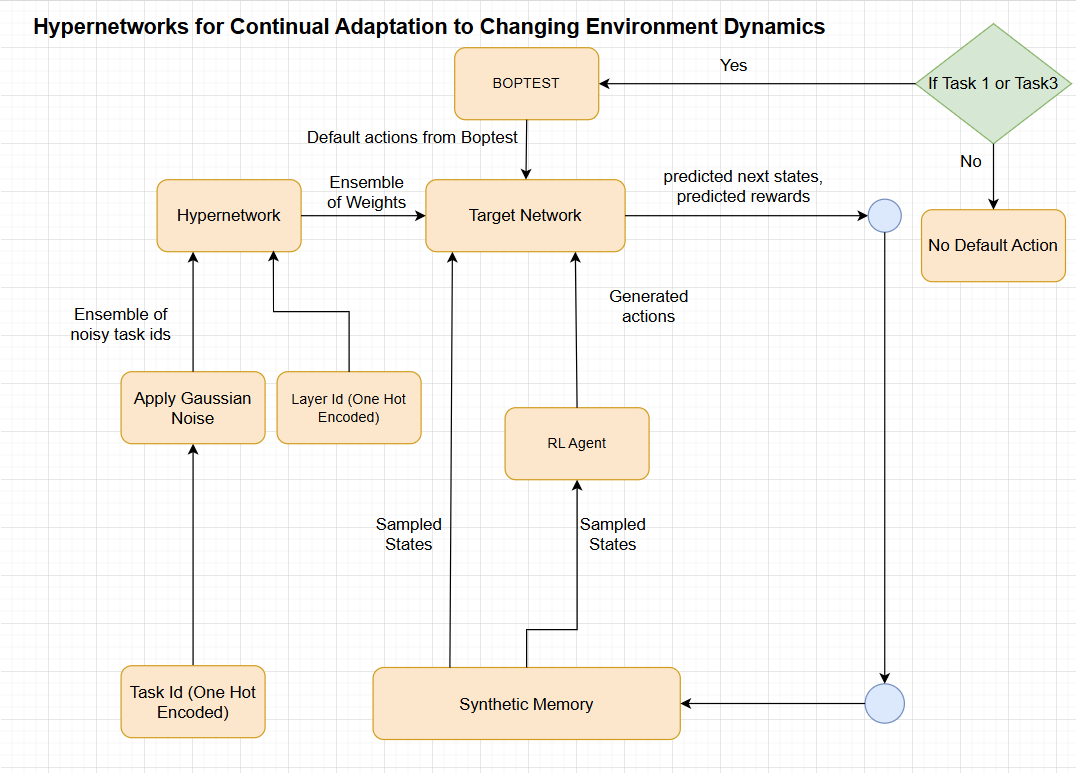}
  \caption{Hypernetwork architecture}
  \label{fig:hypernetwork}
\end{figure}

\clearpage
\vspace{-1em}  
\subsection{Algorithm}
\vspace{-0.5em}  

\begin{algorithm}[ht]
\caption{Continual Model-Based RL with Hypernetwork for Environment Modeling}
\label{alg:hypernet-rl}
\begin{algorithmic}[1]

\State Initialize policy $\pi_{\theta}$
\State Initialize hypernetwork $H_{\varphi}$, target network $T_{\delta}$
\State Initialize real data memory $\mathcal{M}_{\alpha}$, synthetic data memory $\mathcal{M}_{\beta}$
\State Initialize hypernet train memory $\mathcal{M}_{\gamma}$
\State Initialize regularization constant $\alpha$

\For{each task $T_i$ in sequence}
    \State Reinitialize policy $\pi_{\theta}$
    \State Set training iterations $K_i$
    \For{$k = 1$ to $K_i$}
        \State Interact with real environment $T_i$: collect $(s, a, r, s')$ using $\pi_{\theta}$ and store in $\mathcal{M}_{\alpha}, \mathcal{M}_{\gamma}$
        \State Train $H_{\varphi}$ using samples from $\mathcal{M}_{\gamma}$:
        \State \quad $T_{\delta} \gets H_{\varphi}(\texttt{task\_id}, \texttt{layer\_id})$
        \State \quad sample $(s,a,s',r)$ from $\mathcal{M}_{\gamma}$
        \State \quad predict $s', r \gets T_{\delta}(s,a)$
        \State \quad $\text{mse\_loss} \gets \text{MSE(predictions, actual)}$
        \State \quad $\text{regularization} \gets \text{MSE}(T_{\delta,\texttt{old}}, T_{\delta})$
        \State \quad $\text{loss} \gets \text{mse\_loss} + \beta \cdot \text{regularization}$
        \State \quad Update $H_{\varphi}$ by minimizing $\text{loss}$
        \State Generate one-step rollout using $H_{\varphi}$ and $T_{\delta}$:
        \State \quad sample actual state $s$ from $\mathcal{M}_{\alpha}$
        \State \quad take random action $a$ using $\pi_{\theta}$
        \State \quad use $(s,a)$ to generate synthetic $(s',r)$ using $H_{\varphi}$ and $T_{\delta}$
        \State \quad store tuple $(s,a,s',r)$ in $\mathcal{M}_{\beta}$
        \State Update policy $\pi_{\theta}$ using data from $\mathcal{M}_{\alpha}, \mathcal{M}_{\beta}$
    \EndFor
    \State Optionally store $H_{\varphi}$ state or evaluate across tasks
\EndFor

\end{algorithmic}
\end{algorithm}

\subsection*{Implementation Details}

All experiments were conducted on a system with the following specifications:  
\textbf{Processor:} 13th Gen Intel(R) Core(TM) i9-13900 @ 2.00 GHz  
\textbf{Installed RAM:} 32.0 GB 
\textbf{System Type:} 64-bit operating system, x64-based processor  

Our model is implemented in Python using PyTorch. The complete codebase is available at:  
\href{https://github.com/gauthambekal93/hvac_continual_rl}{\texttt{https://github.com/gauthambekal93/hvac\_continual\_rl}}

\clearpage

\subsection*{Complete Flowchart}
\vspace{-1em}  

\begin{figure}[H]
  \centering
  \includegraphics[width=\linewidth, height=0.7\textheight, keepaspectratio]{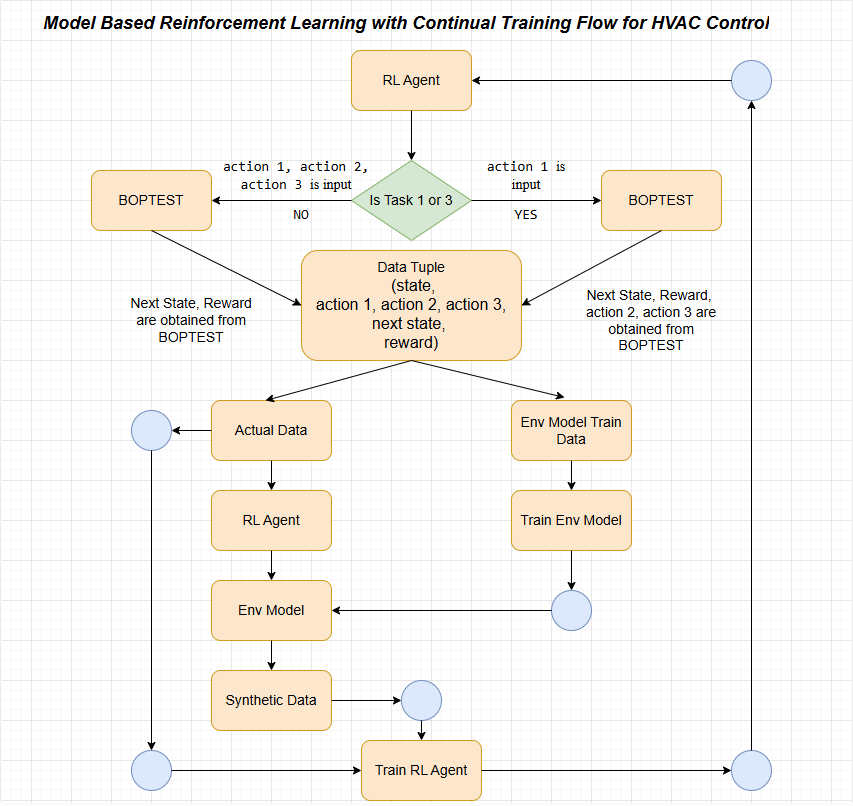}
  \caption{Training Loop}
  \label{fig:trainingloop}
\end{figure}

\vspace{-1em}  

\FloatBarrier  

\subsection{Input Output Mappings}

\subsection*{Model I/O Tables}

\begin{table}[H]
\centering
\caption{Hypernetwork Inputs and Outputs}
\label{tab:hypernet-io}
\begin{tabular}{|l|p{10cm}|}
\hline
\textbf{Component} & \textbf{Description} \\
\hline
Inputs & One-hot encoded Task ID, One-hot encoded Layer ID \\
Outputs & Parameters for Target Model \\
\hline
\end{tabular}
\end{table}

As shown in Table~\ref{tab:hypernet-io}, the hypernet takes one hot encoded values as input for Task ID and Layer ID. For example, task 1 can be represented as [1, 0, 0]  respectively. The outputs are continuous real values, since they represent the parameters of the Target Network.

\vspace{-1em}

\begin{table}[H]
\centering
\caption{Target Network Inputs and Outputs}
\label{tab:targetnet-io}
\begin{tabular}{|l|p{10cm}|}
\hline
\textbf{Component} & \textbf{Description} \\
\hline
Inputs & Real-time Zone Temperature, Dry Bulb Temperature Forecast,  Cooling Setpoint, Heating Setpoint , Action 1, Action 2, Action 3\\
Outputs & Next Real-time Zone Temperature, Reward \\
\hline
\end{tabular}
\end{table}
The Target Network predicts the next zone temperature and reward using the input states and actions. The actions are obtained from RL\_agent and BOPTEST environment depending on task-id.

\vspace{-1em}

\begin{table}[H]
\centering
\caption{RL Agent Inputs and Outputs}
\label{tab:rl-agent-io}
\begin{tabular}{|l|p{12cm}|}
\hline
\textbf{Component} & \textbf{Description} \\
\hline
Inputs & Time, Real-time Zone Temperature, Dry Bulb Temperature Forecast \\
\hline
Outputs &
\textbf{Task 1 or Task 3:} \\
& \quad Action 1 – Heat pump modulating signal \\
& \textbf{Task 2:} \\
& \quad Action 1 – Heat pump modulating signal \\
& \quad Action 2 – Heat pump evaporator fan \\
& \quad Action 3 – Emission circuit pump \\
\hline
\end{tabular}
\end{table}

\gls{sac} model will generate 1 or 3 actions at output based on the task-id.

\subsection{Parameters used}

Table~\ref{tab:hyperparams} lists a 0.99 discount factor and small learning rates for stability, with large buffers and 100-model ensembles to enhance diversity and reduce bias.

\begin{table}[H]
\centering
\caption{Hyperparameter used in experiments}
\label{tab:hyperparams}
\begin{tabular}{|l|c|}
\hline
\textbf{Hyperparameter} & \textbf{Value} \\
\hline
Discount factor & 0.99 \\
Actor learning rate & 0.00005 \\
Critic learning rate & 0.0002 \\
Discount factor ($\gamma$) & 0.99 \\
batch size &  1024 \\
Real data buffer size & 35000 \\
Synthetic data buffer size & 35000 \\
Hypernet learning rate & 0.0001 \\
Hypernet training data buffer size & 4000 \\
Policy update frequency & Every 2 steps \\
Regularization coefficient ($\beta$) & 0.1 \\
No. of time steps per episode ($K_i$) & 1344 \\
No. of synthetic samples per time step & 10 \\
No. of models generated by Hypernet (for bias reduction) & 100 \\
\hline
\end{tabular}
\end{table}

\section{Experimental Results}

\begin{figure}[H]
  \centering
  \begin{minipage}[t]{0.48\textwidth}
    \centering
    \includegraphics[width=\textwidth]{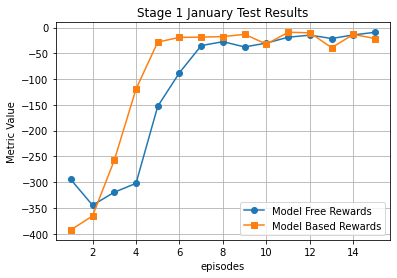}
    \caption{Stage 1 January test results}
    \label{fig:stage1-jan}
  \end{minipage}W
  \hfill
  \begin{minipage}[t]{0.48\textwidth}
    \centering
    \includegraphics[width=\textwidth]{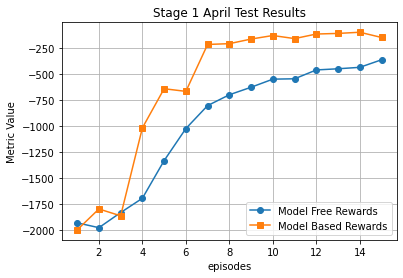}
    \caption{Stage 1 April test results}
    \label{fig:stage1-april}
  \end{minipage}

    \vspace{0.5em} 
  \begin{minipage}[t]{\textwidth}
    \small
    To compare the performance of the system, under different seasonal conditions we depicted Fig.~\ref{fig:stage1-jan} and Fig.~\ref{fig:stage1-april}  which shows the results for January and February respectively.
    For task 1, the actor generates only a heat pump modulating signal as output. The heat pump evaporator fan and emission circuit pump are set to default values and directly obtained from the BOPTEST environment. All three actions are given as input to Hypernet. 
    The left shows results from January testing, and the right shows April performance, illustrating the \gls{rl} agent control performance. This is task 1, which is the pretraining phase. The Hypernet is trained from scratch along with SAC models which are trained from scratch.
    As seen in the literature, \gls{mbrl} performs significantly better than \gls{mfrl} for both the January and April test periods.
    
  \end{minipage}

\end{figure}

\begin{figure}[H]
  \centering
  \begin{minipage}[t]{0.48\textwidth}
    \centering
    \includegraphics[width=\textwidth]{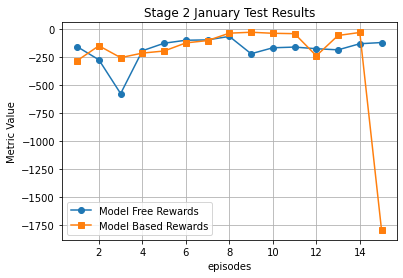}
    \caption{Stage 2 January test results}
    \label{fig:stage2-jan}
  \end{minipage}
  \hfill
  \begin{minipage}[t]{0.48\textwidth}
    \centering
    \includegraphics[width=\textwidth]{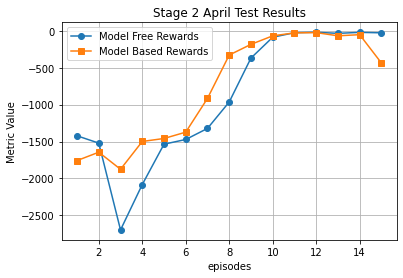}
    \caption{Stage 2 April test results}
    \label{fig:stage2-april}
  \end{minipage}

      \vspace{0.5em} 
  \begin{minipage}[t]{\textwidth}
    \small
     
    In task 2, Fig.~\ref{fig:stage2-jan} and Fig.~\ref{fig:stage2-april} shows the finetuning of the Hypernets after initial training in task 1.
    For task 2, the actor generates three actions, heat pump modulating signal, heat pump evaporator fan and emission circuit pump as output, which are given as input to Hypernet.
    The SAC agent is again trained from scratch for both \gls{mfrl} and \gls{mbrl}.
    However, we utilize transfer learning on the Hypernet which was already trained on task 1.
    We see that for the January and April test periods, the MBRL tends to perform better from early on, however, caution should be taken when training Hypernets since they can have a significant drop in performance due to instabilities. 
  \end{minipage}
  
\end{figure}

\begin{figure}[H]
  \centering
  \begin{minipage}[t]{0.48\textwidth}
    \centering
    \includegraphics[width=\textwidth]{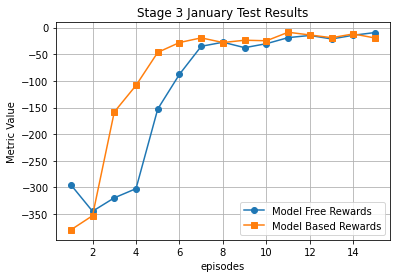}
    \caption{Stage 3 January test results}
    \label{fig:stage3-jan}
  \end{minipage}
  \hfill
  \begin{minipage}[t]{0.48\textwidth}
    \centering
    \includegraphics[width=\textwidth]{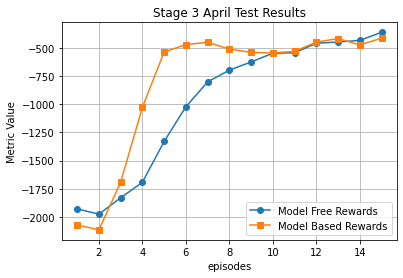}
    \caption{Stage 3 April test results}
    \label{fig:stage3-april}
  \end{minipage}

  \vspace{1em}  

  \begin{minipage}[t]{\textwidth}
    \small
    Fig.~\ref{fig:stage3-jan} and Fig.~\ref{fig:stage3-april} plots show the results after sequential training of the Hypernet for Task 1 and Task 2. The configuration of the action space is the same as in Task 1. The SAC agent is again trained from scratch for both \gls{mfrl} and \gls{mbrl}. Even with minimal retraining of the Hypernet on Task 1, we observe significantly quicker convergence compared to vanilla model-free RL. This demonstrates that the Hypernet retained its ability to accurately model environment dynamics after being trained on Task 2, effectively mitigating catastrophic forgetting with minimal retraining.
  \end{minipage}
\end{figure}

As can be seen from the three sequential tasks it can be seen that, \gls{mbrl} using Hypernetworks can help in faster convergence than \gls{mfrl}. From tasks 1 and 2, it can be observed that Hypernetwork is able to retain the environment dynamics, even after the intermediate training for task 2. However, training Hypernetworks can be challenging, as evidenced by the results in Task 2, where a sudden drop in performance was observed despite previously stable behavior.

\section{Conclusion and Future work}
In this work, we proposed a model-based reinforcement learning framework for HVAC control that leverages a Hypernetwork to learn the dynamics of the environment in tasks with varying action spaces. Our approach enables transfer learning by conditioning dynamics generation on task and layer identifiers, allowing the model to adapt to different control configurations. Experimental results in multiple BOPTEST scenarios demonstrate that our method outperforms model-free RL in terms of sample efficiency.  Additionally, the Hypernetwork shows strong backward transfer capabilities, retaining performance on earlier tasks even after training on new ones, thereby mitigating catastrophic forgetting with minimal retraining.\\

While our results are promising, several directions remain for future investigation:
While we performed our experiments on 3 tasks, it is possible to extend to several sequential tasks and is a part of our ongoing research. Also, we can extend the method to handle multi-zone environments with more complex dynamics and interactions. Finally, validating the approach in real or emulated buildings beyond BOPTEST to assess practical applicability and safety constraints will be a promising direction.


\end{document}